\documentclass[11pt,a4paper]{article}

\usepackage{authblk}

\usepackage{tikz}
\usepackage{amsmath}
\usepackage{array}
\usetikzlibrary{decorations.pathmorphing} % noisy shapes
\usetikzlibrary{fit}          % fitting shapes to coordinates
\usetikzlibrary{backgrounds}  % drawing the background after the foreground
\usetikzlibrary{shapes}
\usetikzlibrary{positioning,calc}
\usepackage{chngcntr}
\usepackage[hyperref]{acl2018}
\usepackage{times}
\usepackage{latexsym}

\usepackage{url}

\aclfinalcopy % Uncomment this line for the final submission
\def\aclpaperid{206} %  Enter the acl Paper ID here

\title{A Simple but Effective Classification Model for \\ Grammatical Error Correction}

\newcommand*\samethanks[1][\value{footnote}]{\footnotemark[#1]}
\author[1\thanks{These two authors contributed equally.}]{Zhu Kaili} 
\author[2\samethanks]{Chuan Wang}
\author[2]{Ruobing Li}
\author[2]{Yang Liu}
\author[1]{Tianlei Hu}
\author[2]{Hui Lin}
\affil[1]{Zhejiang University}
\affil[2]{Shanghai Liulishuo Information Technology Ltd, China}
\affil[1]{\textit {\{kailizhu, htl\}@zju.edu.cn}}
\affil[2]{\textit {\{chuan.wang, ruobing.li, yang.liu, h\}@liulishuo.com}}

\date{}

\begin{document}
\maketitle
\begin{abstract}
  We treat grammatical error correction (GEC) as a classification problem in this study, 
  where for different types of errors, a target word is identified, and the classifier predicts the correct word form
  from a set of possible choices. 
  We propose a novel neural network based feature representation and classification model, trained using large text corpora without human annotations. 
  Specifically we use RNNs with attention to represent both the left and right context of a target word. 
  All feature embeddings are learned jointly in an end-to-end fashion.
  Experimental results show that our novel approach outperforms other classifier methods on the CoNLL-2014 test set ($F_{0.5}$ $45.05\%$). %by a large margin ($F_{0.5}$ $45.05\%$ vs. $41.6\%$). 
  Our model is simple but effective, and is suitable for industrial production.
  % Furthermore, when combined with a machine translation based GEC system, we improved results to $50.16\%$.%, achieving the best result ever reported on this dataset.

\end{abstract}

\section{Introduction}
  %Automatic grammatical error correction (GEC) is an interesting and challenging problem, and is also an essential and useful tool for millions of people who learn English as a second language. These people make a variety of grammar and usage mistakes that are not addressed by standard proofing tools. 
  In recent years, many promising approaches have been proposed for grammatical error correction.
  They can be categorized into two types: classification and machine translation (MT). 
  In the classification approach, for a specific error type, GEC is cast as a classification task (possibly with multiple
  classes), where the class labels represent the correct forms of the words in the sentence. 
  \citet{rozovskaya2014illinois} proposed a classification system CUUI that used different combinations of averaged perceptron, 
  na\"ive Bayes, and pattern-based learning, and was the best classifier method in the CoNLL-2014 shared task \cite{ng2014conll}. 
  \citet{wangdeep} proposed a deep context model that used neural networks to extract the context information of input sentences, 
  without complex feature engineering, and outperformed the previous classifier methods. 
  Another widely used method is based on MT, which aims to translate the incorrect text into correct text directly. 
  \citet{junczys2016phrase} used a statistical MT (SMT) framework Moses and investigated interactions of dense and sparse features, different optimizers, and tuning strategies, and showed good performance.
  \citet{chollampatt2018multilayer} used a neural MT (NMT) system, with a multilayer convolutional encoder-decoder neural network initialized with embeddings that make use of character N-gram information, and achieved outstanding results among all the systems. 

  One problem with the MT based methods is that these models need a large amount of parallel data, where each sentence
  with grammatical errors has its corresponding correct sentence. 
  Annotated training data (with errors labeled) is also needed for standard classifier based approaches.
  However, in this work we use neural networks to learn sentence/context representations for the classifier approach,
  and instead of relying on labeled training data, we generate data for model training from large amount of regular English text.
  Furthermore, we use different attention schemes to capture the dependency among words in the sentences.
  Our simple method does not require elaborated feature engineering for different error types, and can be trained effectively in an end-to-end fashion. 
  Experimental results show that our approach achieves state-of-the-art results on the CoNLL-2014 data among all the classifier methods proposed before. %We have made our codes and trained models public for further imporvements\footnote{https://github.com/zklgame/A-Simple-but-Effective-Classification-Model-for-Grammatical-Error-Correction}.

\section{Classification Task Definition}
We consider five error types: article, preposition, verb form, noun number, and subjective agreement.
A neural classification model is trained for each error type.
Table~\ref{tabel-intro} illustrates the specific classes used for different kinds of errors. 
We treat article error correction as a three-category classification problem: \textit{a/an}, \textit{the} and no article. 
The position where the article can appear should be in front of a noun phrase (a combination of noun words and adjective words). 
For preposition type, we pick 8 most common prepositions: \textit{in}, \textit{to}, \textit{of}, \textit{on}, \textit{by}, \textit{for}, \textit{with} and \textit{about}. 
When the input sentence contains these words, we make a preposition prediction for correction. 
As shown in the table, verb form, noun number, and subjective agreement type can be viewed as 
a three-way, two-way and two-way classification problem respectively.

\begin{table}[!htbp]
\centering

\begin{tabular}{|p{2.2cm}|p{4.5cm}|}
  \hline
  \textbf{Error Type} & \textbf{Classes}\\
  \hline
  article & 0 = a/an, 1 = the, 2 = None\\ 
  \hline
  preposition & label = preposition index\\
  \hline
  verb form & 0 = base form, 1 = gerund or present participle, 2 = past participle\\
  \hline
  noun number & 0 = singular, 1 = plural\\
  \hline
  subjective agreement & 0 = non-3rd person singular present, 1 = 3rd person singular present\\
  \hline
\end{tabular}

\caption{\label{tabel-intro} Classification labels for different error types.}
\end{table}

For each error type, we first use the stanford Corenlp toolkit \cite{manning2014stanford} to 
locate the target words that need to be checked in the given sentence. 
Take sentence (\textit{she eat an apple everyday .}) as an example. 
Its POS tags are (\textit{PRP VBP DT NN NN .}). 
For subject agreement error type, we locate the POS tag \textit{VBP} (verb, non­-3rd person singular present) 
and its corresponding word \textit{eat}, so \textit{eat} is the target word that we check for errors.
Our model predicts a class label that stands for the relative form of the target word, 
and thus we can get the corresponding predicted word.  
For target word \textit{eat}, if the predicted class label is $1$, then the corresponding predicted word is \textit{eats}. 
If the final predicted word is different from the original target word, then the original word will be marked as a mistake 
and be replaced by the predicted word for correction. 
For the given input sentence, we apply the five models corresponding to the five error types in the following order: 
verb form, noun number, article, preposition and subjective agreement.

\section{Neural Model}
%For a certain error type, the corresponding model learns an embedding function for variable-length sentential contexts around the target word to predict the correct word. Our model firstly uses GloVe \cite{pennington2014glove} to initialize the embedding of each input word and sends the word embeddings to two gated recurrent unit (GRU) \cite{cho2014properties} to produce left and right context states with attention mechanism. Then we concatenate both states and feed it into a MLP to get the predicted label.

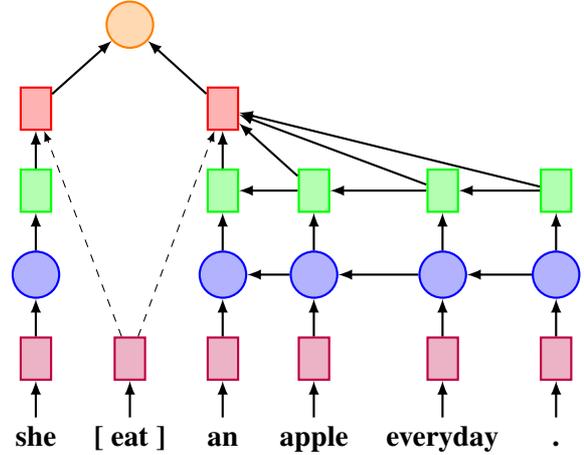
\begin{figure}[!htbp]
\centering

% \definecolor{color1}{RGB}{85,162,225}

\tikzstyle{input}=[rectangle,
                  thick,
                  minimum width=0.7cm,
                  minimum height=0.4cm,
                  inner sep=0.1cm
                  ]
\tikzstyle{embedding}=[rectangle,
                      thick,
                      minimum width=0.4cm,
                      minimum height=0.5cm,
                      draw=purple,
                      fill=purple!30]
\tikzstyle{embedding_2}=[rectangle,
                      thick,
                      minimum width=0.4cm,
                      minimum height=0.5cm,
                      draw=purple,
                      fill=purple!30]
\tikzstyle{rnn}=[circle,
                      thick,
                      minimum size=0.4cm,
                      draw=blue,
                      fill=blue!30]
\tikzstyle{rnn_hidden}=[circle,
                      thick,
                      minimum size=0.4cm,
                      ]
\tikzstyle{output}=[rectangle,
                      thick,
                      minimum width=0.4cm,
                      minimum height=0.5cm,
                      draw=green,
                      fill=green!30]
\tikzstyle{output_hidden}=[rectangle,
                      thick,
                      minimum width=0.4cm,
                      minimum height=0.5cm,
                      ]
\tikzstyle{state}=[rectangle,
                      thick,
                      minimum width=0.4cm,
                      minimum height=0.5cm,
                      draw=red,
                      fill=red!30]
\tikzstyle{state_hidden}=[rectangle,
                      thick,
                      minimum width=0.4cm,
                      minimum height=0.5cm,
                      ]
\tikzstyle{mlp}=[circle,
                  thick,
                  minimum size=0.5cm,
                  draw=orange,
                  fill=orange!30]
\tikzstyle{mlp_hidden}=[circle,
                  thick,
                  minimum size=0.5cm,
                  ]
\tikzstyle{background}=[rectangle,
                        %fill=gray!30,
                        inner sep=0.0cm,
                        rounded corners=3mm]

\begin{tikzpicture}[>=latex,text height=1.5ex,text depth=0.25ex]
    
    \matrix[row sep=0.5cm,column sep=0.3cm](inmtx) {
        \node (mlp_h1)      [mlp_hidden] {}; &
        \node (mlp)         [mlp] {}; &
        \node (mlp_h2)      [mlp_hidden] {}; &
        \node (mlp_h3)      [mlp_hidden] {}; &
        \node (mlp_h4)      [mlp_hidden] {}; &
        \node (mlp_h5)      [mlp_hidden] {}; &
        \\
        \node (state_l)      [state] {}; &
        \node (state_h1)     [state_hidden] {}; &
        \node (state_r)      [state] {}; &
        \node (state_h2)     [state_hidden] {}; &
        \node (state_h3)     [state_hidden] {}; &
        \node (state_h4)     [state_hidden] {}; &
        \\
        \node (She_o)      [output] {}; &
        \node (eat_o)      [output_hidden] {}; &
        \node (an_o)       [output] {}; &
        \node (apple_o)    [output] {}; &
        \node (everyday_o) [output] {}; &
        \node (eos_o)      [output] {}; &
        \\
        \node (She_r)      [rnn] {}; &
        \node (eat_r)      [rnn_hidden] {}; &
        \node (an_r)       [rnn] {}; &
        \node (apple_r)    [rnn] {}; &
        \node (everyday_r) [rnn] {}; &
        \node (eos_r)      [rnn] {}; &
        \\
        % \node (She_e) [embedding] {$\mathbf{e}_1$}; &
        % \node (eat_e) [embedding] {$\mathbf{e}_2$}; &
        % \node (an_e) [embedding] {$\mathbf{e}_3$}; &
        % \node (apple_e) [embedding] {$\mathbf{e}_4$}; &
        % \node (everyday_e) [embedding] {$\mathbf{e}_5$}; &
        % \node (eos_e) [embedding] {$\mathbf{e}_6$}; &
        \node (She_e) [embedding_2] {}; &
        \node (eat_e) [embedding_2] {}; &
        \node (an_e) [embedding_2] {}; &
        \node (apple_e) [embedding_2] {}; &
        \node (everyday_e) [embedding_2] {}; &
        \node (eos_e) [embedding_2] {}; &
        \\
        \node (She)      [input] {$\textbf{she}$}; &
        \node (eat)      [input] {$\textbf{[ eat ]}$}; &
        \node (an)       [input] {$\textbf{an}$}; &
        \node (apple)    [input] {$\textbf{apple}$}; &
        \node (everyday) [input] {$\textbf{everyday}$}; &
        \node (eos)      [input] {$\textbf{.}$}; &
        \\
    };

    \path[->]
        (She) edge[thick] (She_e)
        (eat) edge[thick] (eat_e)
        (an) edge[thick] (an_e)
        (apple) edge[thick] (apple_e)
        (everyday) edge[thick] (everyday_e)
        (eos) edge[thick] (eos_e)

        (She_e) edge[thick] (She_r)
        % (eat_e) edge[dashed] (eat_r)
        (an_e) edge[thick] (an_r)
        (apple_e) edge[thick] (apple_r)
        (everyday_e) edge[thick] (everyday_r)
        (eos_e) edge[thick] (eos_r)

        (She_r) edge[thick] (She_o)
        % (eat_r) edge[dashed] (eat_o)
        (an_r) edge[thick] (an_o)
        (apple_r) edge[thick] (apple_o)
        (everyday_r) edge[thick] (everyday_o)
        (eos_r) edge[thick] (eos_o)

        % (She_r) edge[dashed] (eat_r)
        (apple_r) edge[thick] (an_r)
        (everyday_r) edge[thick] (apple_r)
        (eos_r) edge[thick] (everyday_r)
        % (She_o) edge[dashed] (eat_o)
        (apple_o) edge[thick] (an_o)
        (everyday_o) edge[thick] (apple_o)
        (eos_o) edge[thick] (everyday_o)

        (She_o) edge[thick] (state_l)
        % (eat_o) edge[dashed] (state_l)
        (an_o) edge[thick] (state_r)
        (apple_o) edge[thick] (state_r)
        (everyday_o) edge[thick] (state_r)
        (eos_o) edge[thick] (state_r)
        (eat_e) edge[dashed] (state_l)
        (eat_e) edge[dashed] (state_r)

        (state_l) edge[thick] (mlp)
        (state_r) edge[thick] (mlp)
        ;

   % \begin{pgfonlayer}{background}
   %      \node [background,
   %              fit=(She_e) (eos_e),
   %              label=left:embeddings] {};
   %      \node [background,
   %              fit=(She_r) (eos_r),
   %              label=left:rnn layer] {};
   %      \node [background,
   %              fit=(She_o) (eos_o),
   %              label=left:rnn outputs] {};
   %      \node [background,
   %              fit=(state_l) (state_h4),
   %              label=left:attention states] {};
   %      \node [background,
   %              fit=(mlp) (mlp),
   %              label=left:mlp] {};
   %  \end{pgfonlayer}

\end{tikzpicture}

\caption{\label{model-overview} Model Architecture. From bottom to top are embeddings, RNN layer, RNN outputs, attention states and MLP.}

\end{figure}

Figure~\ref{model-overview} illustrates our model architecture. 
We uses GloVe \cite{pennington2014glove} to initialize the embedding of each input word. 
After we identify the target word for an error type (e.g., \textit{eat} in the example sentence), 
we split the sentence into two parts, its left context (i.e., \textit{she}), and its right context (i.e., \textit{an apple everyday .}),
and use gated recurrent unit (GRU) \cite{cho2014properties} layer to represent them respectively. 
%For most error types, the information around the target word is enough to predict the target word form correctly, and what the target word really is is not so important, so we can omit it. But for the noun number and verb form error type, what form the target word should be is closely related to itself, so in these cases, we should add the base form of the target word to calculate the attention-based state for GEC. 
That is, given an embedded sentence $e_{1:n}$ and a target word $w_{i}$, we have the left GRU outputs $lo_{1:i-1}$, and the right GRU outputs $ro_{i+1:n}$. 

We propose two kinds of attention mechanism to better represent the context and the target word.
The difference between them is whether to use the target word information. 

The first attention uses context words only, without the target word.
This is to model the inner connection among the input context words. 
The following equations show how attention is calculated for the left GRU outputs $lo_{1:i-1}$.
\begin{align}
\begin{split}
score(lo_{t}) &= lo_{t}^TW_{a}lo_{i-1} \\
a(t) &= \frac{exp(score(lo_{t}))}{\sum_{j=1}^{i-1}exp(score(lo_{j}))} \\
lstate &= (\sum_{t=1}^{i-1}a(t)lo_{t}) \oplus lo_{i-1}
\end{split}
\end{align}
where $W_{a}$ is a matrix.
The final left context is then represented as a weighted average vector, concatenated with the last GRU output. 
The formula for right context state is similar, except that $ro_{i+1}$ is the last GRU output,
because we feed the right context words into the GRU layer from right to left.

The second kind of attention uses the target word to model the relationship between it and its context.
The following equations are for the left context.
\begin{align}
\begin{split}
score(lo_{t}) &= lo_{t}^TW_{b}\hat{e_{i}} \\
a(t) &= \frac{exp(score(lo_{t}))}{\sum_{j=1}^{i-1}exp(score(lo_{j}))} \\
lstate &= \sum_{t=1}^{i-1}a(t)lo_{t} \oplus \hat{e_{i}} \oplus lo_{i-1}
\end{split}
\end{align}
where $W_{b}$ is a matrix, and $\hat{e_{i}}$ is the embedding of the base form of the target word. 
The left context state is the weighted average vector, concatenated with the target word embedding, and the last GRU output. 
Again similar attention is applied to the right context. 
We only use this target word dependent attention to noun number and verb form error types,
because when deciding the correct word form for these two types, the interaction between the word itself and its contexts is crucial;
whereas for the other three types, the target word information is not necessary or not correct to be useful,
so we only use the first attention for them. 
For preposition error type, the target word itself may be wrong 
(e.g., the target word is \textit{of}, and the right word may be \textit{to}). 
For subjective agreement or article error types, the information about the target word (the verb, article location) 
is not needed to predict the correct form. 

After calculating the left and right states, we concatenate them and feed it to a multilayer perceptron (MLP). 
At the last layer of the MLP, we use a softmax function to calculate the probability of each class for an error type:
\begin{align}
\begin{split}
L(x) &= Wx + b \\
MLP(x) &= softmax(L(ReLU(L(x))))
\end{split}
\end{align}
where $ReLU$ is the Rectified Linear Unit activation function \cite{nair2010rectified}, 
and $L(x)$ is a fully connected linear operation. The output of our model is the label with the highest probability. 

To train the model, we use cross entropy loss: 
\begin{align}
  loss = \frac{1}{n}\sum_{i=1}^{n}y_{i}\log{\hat{y_{i}}}
\end{align}
where $n$ is the number of training samples, $\hat{y_{i}}$ is the predicted label, 
and $y_{i}$ is the true label.

\section{Experiments}

\subsection{Setup}
We use the wiki dump\footnote{https://dumps.wikimedia.org/enwiki/} 
and COCA\footnote{https://corpus.byu.edu/coca/} corpora to generate training data for five grammatical types separately. 
For example, for subjective agreement type, we locate the VBP and VBZ POS tags 
and the corresponding target word in a sentence, and then the word and its left/right context can be used for model training. 
All the input text is lowercased. 
The vocabulary is made up of the most $40K$ common words in the corpora,
and all tokens that are not in the vocabulary are represented as a single \textit{unk} token. 

We use part of the CoNLL-2014 training dataset as our validation dataset.
We evaluate our model on the CoNLL-2014 test set, and report $F_{0.5}$ result that is a standard metric for
this error correction task. 
$F_{0.5}$ combines precision (P) and recall (R), while emphasizing precision twice as much as recall, 
since accurate feedback is often more important than coverage in error correction.

% put it to appendix because of the page limit
% \input{inserts/model_settings}

We trained the classifiers separately for each error type.
Some important model settings are provided in appendix (Table~\ref{tabel-model-settings}).

After we obtain the model's prediction, we set a threshold (e.g., 0.9) for each error type in the final error correction process.
If the probability for the predicted label is higher than the threshold and the corresponding word form is different 
from that of the original target word, we use the predicted one as the correction.

\subsection{Results}

\subsubsection{Type-specific Results}
\label{Type-specific Results}

Table~\ref{tabel-subresults-attn} shows the results of our model for different error types, with and without attention. 
For all the error types, we can see that using attention achieves better performance,
suggesting the effectiveness of modeling the interaction among words in the context,
or between the target word and other words.

Table~\ref{tabel-subresults} compares our model, CUUI, and the deep context model. 
The deep context model uses only the wiki dump data and the base form of the target noun word as extra context information, 
without attention. 
Our model achieves the highest $F_{0.5}$ scores for all the error types except the preposition one that has a small performance
degradation.
Some example corrections from our model are provided in appendix (Table~\ref{tabel-samples}).

% put it here to adjust page display
\begin{table}[!htbp]
\centering

\begin{tabular}{|c|c|c|c|}
  \hline
  \textbf{Error Type} & \textbf{Baseline} & \textbf{Attention} \\
  \hline
  article & 46.80 & \textbf{48.83}\\ 
  \hline
  preposition & 17.44 & \textbf{18.57}\\ 
  \hline
  verb form & 27.71 & \textbf{33.42}\\ 
  \hline
  noun number & 25.24 & \textbf{50.30}\\ 
  \hline
  subjective agreement & 52.95 & \textbf{57.79}\\ 
  \hline
  % \hline
  % \textbf{Error Type} & \textbf{Baseline} & \textbf{Attention} & $+/-$ \\
  % \hline
  % article & 46.80 & 48.83 & +2.03\\ 
  % \hline
  % preposition & 17.44 & 18.57 & +1.13\\ 
  % \hline
  % verb form & 27.71 & 33.42 & +5.71\\ 
  % \hline
  % noun number & 25.24 & 50.30 & +25.06\\ 
  % \hline
  % subjective agreement & 52.95 & 57.79 & +4.84\\ 
  % \hline
\end{tabular}

\caption{\label{tabel-subresults-attn} Results of our neural classification model, with and without attention,
on CoNLL-2014 data (based on the combination of two annotators without alternative answers). }
\end{table}
\begin{table}[!htbp]
\centering

\begin{tabular}{|p{1.8cm}<{\centering}|p{0.85cm}<{\centering}|p{0.85cm}<{\centering}|p{0.85cm}<{\centering}|p{0.85cm}<{\centering}|}
  \hline
  \textbf{Error Type} & $\textbf{A}$ & \textbf{B} & \textbf{Ours} & $+/-$ \\
  \hline
  article & 33.7 & 42.1 & \textbf{48.83} & +6.7 \\ 
  \hline
  preposition & 19.0 & \textbf{19.1} & 18.57 & -0.5 \\ 
  \hline
  verb form & 19.2 & 15.3 & \textbf{33.42} & +14.2\\ 
  \hline
  noun number & 41.0 & 42.4 & \textbf{50.30} & +7.9 \\ 
  \hline
  subjective agreement & 49.3 & 49.9 & \textbf{57.79} & +7.9\\ 
  \hline
\end{tabular}

\caption{\label{tabel-subresults} Results of our model, in comparison with the best classifier CUUI (A) in CoNLL-2014 
and the deep context model (B). 
The last column means the difference with the previous highest score. 
Again the data is the combination of two annotators without alternative answers in CoNLL-2014. }
\end{table}

\subsubsection{Overall Results}

Finally we fix the mechanical errors (punctuation, spelling and capitalization errors) 
using existing resources and rule-based methods similar to \cite{rozovskaya2016grammatical},
since these errors are different from the grammatical mistakes and not specific to GEC. 
After that, our model corrects five type errors in order. 
In addition, we also combine our model with the public SMT system from \cite{rozovskaya2016grammatical} 
to build a hybrid system by letting our model and the SMT take turns to correct grammatical errors until there is no change 
for the input sentence. 
Table~\ref{tabel-results} presents the results of ours in comparison to several previously published best results 
on the CoNLL-2014 shared task data.

Our neural classification model outperforms the CUUI system and the deep context model, and has similar performance as \cite{ji2017nested}, the first best fully neural MT method. 
%\citet{rozovskaya2016grammatical} explored key strengths of both the classifier approach and the MT approach for GEC, 
%and combined the strengths of both approaches to build a powerful hybrid system. 
Our hybrid method consisting of our neural classification model and the public SMT system (that is, replacing the classifier method in \cite{rozovskaya2016grammatical}) has a better performance with a $50.16$ $F_{0.5}$ score. \cite{chollampatt2018multilayer} achieved the best performance as we can see, but our model is a single model without re-scoring using edit operation and language model features, and is more suitable for industrial production.

% put it here to adjust page display
\begin{table}[!htbp]
\centering

\begin{tabular}{|p{3.2cm}<{\centering}|p{0.8cm}<{\centering}|p{0.8cm}<{\centering}|p{0.8cm}<{\centering}|}
  \hline
  \textbf{System} & \textbf{P} & \textbf{R} & \textbf{$F_{0.5}$} \\
  \hline
  CUUI & 41.78 & 24.88 & 36.79 \\
  % \hline
  % CoNLL-2014 top-1 & 39.71 & 30.10  & 37.33 \\
  \hline
  \textbf{The public SMT} & 66.02 & 15.11 & 39.44 \\
  \hline
  \cite{wangdeep} & 54.5 & 21.3 & 41.6\\
  \hline
  \textbf{Our neural model} & 58.18 & 23.68 & \textbf{45.05} \\
  \hline
  \cite{ji2017nested} & N/A & N/A& 45.15\\
  \hline
  \cite{rozovskaya2016grammatical} & 60.17 & 25.64  & 47.40 \\
  \hline
  % \cite{junczys2016phrase} & 61.27 & 27.98 & 49.49\\
  % \hline
  \textbf{Our system} & 59.36 & 30.97 & \textbf{50.16} \\
  \hline
  \cite{chollampatt2018multilayer} & 65.49 & 33.14 & 54.79 \\
  \hline
\end{tabular}

\caption{\label{tabel-results} Overall performance of our model and other systems.}
\end{table}

%The top-2 system in CoNLL-2014 task (CUUI in Table~\ref{tabel-subresults}) only builds classifiers for specific error types and does not attempt to tackle the whole range of errors. The top-1 system in CoNLL-2014 task is a hybrid system combining rules and machine translation methods. The deep context model proposed by \cite{wangdeep} has been compared in Subsection~\ref{Type-specific Results}. Our model makes full use of the deep GRU-based neural model and attention mechanism, outperforms these systems significantly. 

%\citet{ji2017nested} proposes attention on neural machine translation and beats other fully neural systems with a $45.15$ $F_{0.5}$ score. In \cite{rozovskaya2016grammatical}, the authors explore key strengths of both the classifier approach and the MT approach for GEC, and combine the strengths of both approaches to build a powerful hybrid system. 
%\citet{junczys2016phrase} integrates discriminative training toward the task-specific evaluation function, a rich set of features, and multiple large language models, and reports the highest performance to date on the task of $49.5$ in $F_{0.5}$ score on the CoNLL-14 test set. 
%Now, our hybrid system that consists of our neural models and the public trained SMT, makes full use of the strengths of both neural classification and machine translation, thus makes a better performance on the GEC task with a $50.16$ $F_{0.5}$ score.

\section{Conclusions}
We propose a neural classification model to learn context representation of sentences for grammatical error correction. 
Attention mechanisms are designed to properly model characteristics of different error types.
Compared to the traditional classifier methods, our approach does not need complex feature engineering, 
 the context representation is learned jointly with classification in an end-to-end fashion, 
and we can effectively utilize enormous and easy-to-get native data. 
This method outperforms other classifier approaches, and is more suitable for industrial production compared with the state-of-the-arts.

%\bibliography{acl2018}
\bibliography{acl2018}
\bibliographystyle{acl_natbib}

\appendix
\counterwithin{table}{section}
\section{Supplemental Material}
\begin{table*}[htp]
\centering

\begin{tabular}{|c|c|c|c|c|c|c|}
  \hline
  \textbf{Error Type} & \textbf{Attention Type} & \textbf{Optimizer} & \textbf{LR} & \textbf{GRU} & \textbf{Threshold}\\
  \hline
  article & first & SGD & 0.08 & 128 & 0.9\\ 
  \hline
  preposition & first & SGD & 0.08 & 128 & 0.85\\ 
  \hline
  verb form & second & Adam & 0.001 & 256 & 0.9\\ 
  \hline
  noun number & second & Adam & 0.001 & 256 & 0.9\\ 
  \hline
  subjective agreement & first & SGD & 0.08 & 256 & 0.9\\ 
  \hline
\end{tabular}

\caption{\label{tabel-model-settings} Important model settings of each error type. SGD means stochastic gradient descent algorithm, Adam means the algorithm from \cite{Kingma2014Adam}, LR means learning rate, GRU means GRU hidden size. Other parameters are the same for all types: the word embedding size is 300, MLP hidden size is 512.}
\end{table*}
\begin{table*}[!htbp]
\centering

\begin{tabular}{l|p{6.5cm}|p{6.5cm}}
  \textbf{No.} & \textbf{Original} & \textbf{Proposed} \\
  \hline
  1. & then how does car come into being ... & then how does \underline{the} car come into being ...\\ 
  \hline
  2. & especially for \underline{the} young people without marriage & especially for young people without marriage \\ 
  \hline
  3. & \underline{for} the case of marriage, people should be honest. & \underline{in} the case of marriage, people should be honest.\\ 
  \hline
  4. & ... negative impacts \underline{to} the family & ... negative impacts \underline{on} the family\\ 
  \hline
  5. & he might end up \underline{dishearten} his family. & he might end up \underline{disheartening} his family.\\ 
  \hline
  6. & it will just \underline{adding} on their misery. & it will just \underline{add} on their misery.\\ 
  \hline
  7. & ... be honest with his or her \underline{feeling}. & ... be honest with his or her \underline{feelings}.\\ 
  \hline
  8. & ... after realising his or her \underline{conditions}. & ... after realising his or her \underline{condition}.\\ 
  \hline
  9. & the popularity of social media sites \underline{have} made ... & the popularity of social media sites \underline{has} made ...\\ 
  \hline
  10. & these skills are important to know, but \underline{is} difficult ... & these skills are important to know, but \underline{are} difficult ...\\ 
\end{tabular}

\caption{\label{tabel-samples} Examples of our model corrections. Article errors are demonstrated in the 1st and 2nd sentences. Preposition errors are corrected in 3rd and 4th sentences. The 5th and 6th sentences show that verb form errors can be corrected. Even though the surrounding words are similar in 7th and 8th sentences, our model still successfully corrects these noun number errors. And even though the subjects are not near the verbs, errors are still corrected in the 9th and 10th sentences.}
\end{table*}

\end{document}